\documentclass[journal]{IEEEtran}

\ifCLASSINFOpdf
\usepackage[pdftex]{graphicx}

\else
\fi

\usepackage{amsmath}
\usepackage{amsfonts}
\usepackage{graphicx,verbatim}
\usepackage{tabularx}
\usepackage{multirow}
\usepackage{booktabs}
\usepackage{amssymb}
\usepackage{float} 

\usepackage{hyperref} 
\usepackage{xcolor}

\newcommand{\code}[1]{\texttt{#1}}
\DeclareSymbolFont{bbold}{U}{bbold}{m}{n}
\DeclareSymbolFontAlphabet{\mathbbold}{bbold}

\begin{document}
%
\title{Point Tracking in Surgery--The 2025 Surgical Tattoos in Infrared Challenge (STIRC2025)}
%
%
%

\author{Adam Schmidt$^{1}$, 
    Mert Asim Karaoglu$^{2, 3}$, 
    Zijian Wu$^{4}$,
    Jiaming Zhang$^{5}$,
    Yuxin Chen$^{4}$,
    Tim Salcudean$^{4}$
    Ho-Gun Ha$^{6}$, 
    Minkang Jang$^{6}$,
    Kyungmin Jung$^{6}$, 
    Ihsan Ullah$^{6}$, 
    Hyunki Lee$^{6}$,
    Suresh Guttikonda$^{7}$,
    Sarah Latus$^{7}$,
    Alexander Schlaefer$^{7, 8}$,
    Xinkai Zhao$^{9}$, 
    Yuichiro Hayashi$^{9}$, 
    Masahiro Oda$^{9, 10}$, 
    Takayuki Kitasaka$^{11}$, 
    Kensaku Mori$^{9, 10, 12}$,
    Peng Liu$^{13, 14}$,
    Chenyang Li$^{13, 15}$,
    Stefanie Speidel$^{13, 14, 15, 16}$,
    Aoife Gardiner$^{17}$, 
    Agostino Stilli$^{17}$, 
    Danail Stoyanov$^{17}$, 
    Francisco Vasconcelos$^{17}$,
    Anwesa Choudhuri$^{18}$, 
    Meng Zheng$^{18}$,
    Zhongpai Gao$^{18}$, 
    Benjamin Planche$^{18}$, 
    Van Nguyen Nguyen$^{18}$,
    Terrence Chen$^{18}$,
    Ziyan Wu$^{18}$,
    Alexander Ladikos$^{2}$,
    Omid Mohareri$^{1}$

\thanks{$^1$ Intuitive Surgical Inc., Sunnyvale, USA,
    $^2$ ImFusion GmbH, Munich, Germany,
    $^3$ Technical University of Munich, Munich, Germany,
    $^4$ University of British Columbia, Vancouver, Canada,
    $^5$ Johns Hopkins University, Baltimore, USA,
    $^6$ Daegu Gyeongbuk Institute of Science and Technology, Daegu, South Korea,
    $^7$ Hamburg University of Technology, Hamburg, Germany,
    $^8$ SustAInLivWork Center of Excellence, Kaunas, Lithuania,
    $^9$ Graduate School of Informatics, Nagoya University, Nagoya, Japan,
    $^{10}$ Information Technology Center, Nagoya University, Nagoya, Japan,
    $^{11}$ Department of Information Science, Aichi Institute of Technology, Aichi, Japan,
    $^{12}$ Research Center for Medical Bigdata, National Institute of Informatics, Tokyo, Japan,
    $^{13}$ Department of Translational Surgical Oncology, National Center for Tumor Diseases (NCT), NCT/UCC Dresden, a partnership between DKFZ, Faculty of Medicine and University Hospital Carl Gustav Carus, TUD Dresden University of Technology, and Helmholtz-Zentrum Dresden-Rossendorf (HZDR), Germany,
    $^{14}$ German Cancer Research Center (DKFZ), Heidelberg, Germany
    $^{15}$ Faculty of Medicine and University Hospital Carl Gustav Carus, TUD Dresden University of Technology, Dresden, Germany
    $^{16}$ Centre for Tactile Internet with Human-in-the-Loop, TU Dresden, Dresden, Saxony, Germany
    $^{17}$ Hawkes Institute, University College London, London, UK,
    $^{18}$ United Imaging Intelligence, Boston, MA, USA. 
    }
}

%
%

\markboth{Point Tracking in Surgery--The 2025 Surgical Tattoos in Infrared (STIR) Challenge}%
{Point Tracking in Surgery--The 2025 Surgical Tattoos in Infrared (STIR) Challenge}
%

\maketitle

\begin{abstract}
Point tracking in surgery is crucial to enable applications in downstream tasks such as segmentation, 3D reconstruction, virtual tissue landmarking, autonomous probe-based scanning, and subtask autonomy.
This paper introduces the 2025 iteration of a point tracking challenge to address this, wherein participants submit their algorithms for quantification.
Their algorithms are evaluated using a dataset named surgical tattoos in infrared (STIR), with the challenge named the STIR Challenge 2025 (STIRC2025).
The STIR Challenge 2025 comprises two quantitative components: accuracy and efficiency.
The accuracy component tests the accuracy of algorithms on \em{in vivo} and \em{ex vivo} sequences.
The efficiency component tests algorithm inference latency.
The challenge was conducted as a part of MICCAI EndoVis 2025, and seven teams participated in this challenge.
In this paper we summarize the challenge results and participant methods.
The challenge dataset is available at: \url{https://zenodo.org/records/20191078}, and the code for baseline models and metrics calculation is available here: \url{https://github.com/athaddius/STIRMetrics}
\end{abstract}

\begin{IEEEkeywords}
Endoscopy, Point Tracking, Deformable, Tissue Tracking, Challenge
\end{IEEEkeywords}

\IEEEpeerreviewmaketitle

\section{Introduction}

\IEEEPARstart{T}{he} 2025 STIR challenge is designed to help improve tracking and reconstruction methods in surgery.
Knowledge of tissue motion and location is critical to enable many tasks in medical computer vision~\cite{schmidtTrackingMappingMedical2024a}.
Enabling accurate point tracking is essential for automated dexterity~\cite{kamAutonomousSystemVaginal2023}, autonomous scanning~\cite{zhan2020autonomous}, and virtual landmarking.
Point tracking methods could also benefit foundation models, where physical priors can be incorporated into pretraining.
The data used in the challenge comprises 32 sequences with each sequence having 7 points on average.

\begin{figure*}[t]
	\centering
	\includegraphics[width=\textwidth]{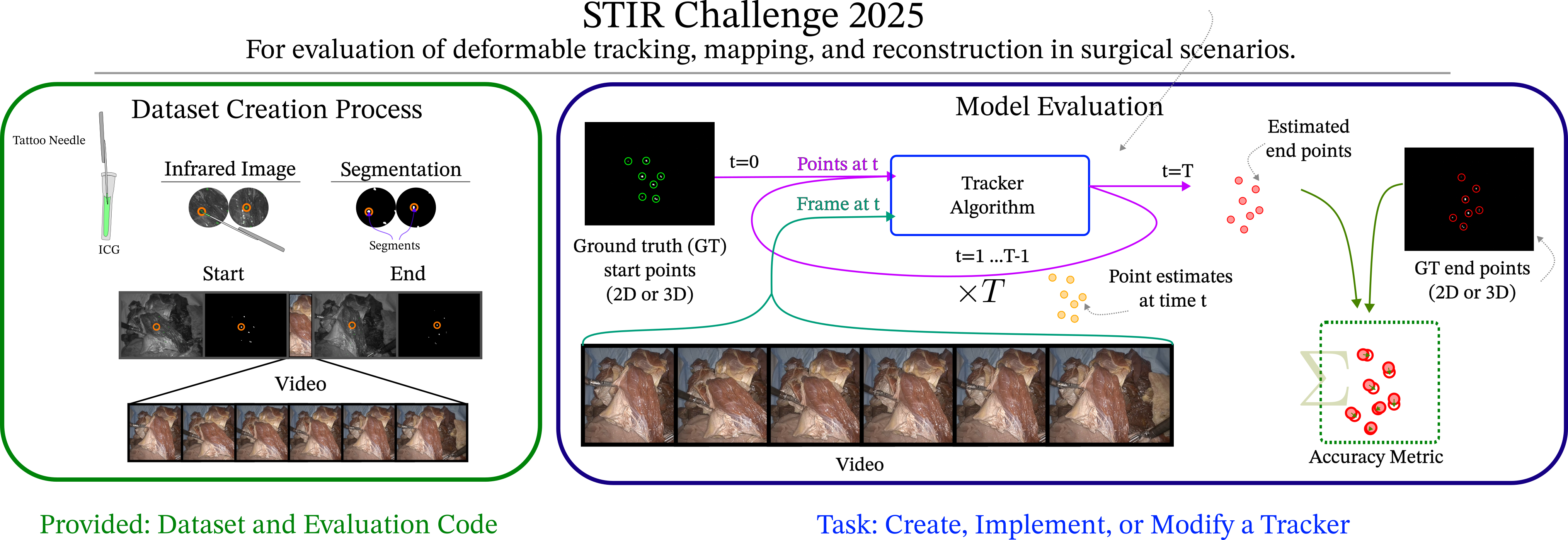}
	\caption{This figure describes the STIR Challenge 2025. Participants submit their algorithms in a docker container. The algorithm receives a video and a list of start points from each sequence in the dataset. Participants use their tracker to estimate the motion of a set of points for every frame in a video. Videos are provided in a streaming manner. The final estimates are then compared to the ground truth labels at the end of the video (Section~\ref{sec:dataandannotation}). The errors are then averaged across all points to obtain the final metrics in 2D or 3D (Section~\ref{sec:metrics}). Latency is calculated alongside the inference for those who participated in the efficiency component of the challenge.}
	\label{fig:challengeexample}
\end{figure*}

In this paper, we will first provide a brief clarification of the challenge dataset compared to the original STIR dataset (STIROrig) and the 2024 Challenge dataset (STIRC2024) in~\ref{sec:diffs}, followed by a non-exhaustive summary of datasets that we see as useful to tracking in Section~\ref{sec:datasets}.
Then, we will describe the dataset format and annotation protocol for the challenge in Section~\ref{sec:dataandannotation}.
Afterwards, we describe the metrics we calculate as part of the challenge in Section~\ref{sec:metrics}.
We then summarize all submissions received in Section~\ref{sec:submissions_and_baselines}, and their results in Section~\ref{sec:results}.
We provide a brief discussion of the results and challenge organization in Section~\ref{sec:discussion}, and finally conclude in Section~\ref{sec:conclusion}.
For a high-level overview of the challenge, refer to Fig.~\ref{fig:challengeexample}.

\subsection{STIR Challenge Data (STIRC2024) vs STIR Original (STIROrig)}
\label{sec:diffs}
Here, we will explain the differences between the STIR Challenge 2025 (STIRC2025), the STIR Challenge 2024 (STIRC2024), and the original STIR dataset.
The original STIR dataset (STIROrig) is a dataset that is publicly released and usable for test, validation, or training (available at: \url{https://ieee-dataport.org/open-access/stir-surgical-tattoos-infrared}).
This dataset is released as a way to validate, test, design, and evaluate algorithms~\cite{schmidtSurgicalTattoosInfrared2024}.
STIROrig remains useful for this exact purpose, in addition to being larger than the challenge dataset.
We note that STIROrig has less outlier and error filtering than the challenge datasets, but is much larger.
The STIR Challenge 2024 dataset (STIRC2024) is a similar dataset that was witheld from the initial STIROrig release in order to enable proper evaluation without the risk of participants fine-tuning or overfitting to already released data.
STIRC2024 has additional filtering and removal of noisy labels, and can be used for fine grained evaluation and testing.
STIRC2024 is available at \url{https://zenodo.org/records/14803158}.
The STIR Challenge 2025 (STIRC2025) dataset was filtered in the same way as STIRC2024, and has no intersection with either of the prior datasets.
Post-challenge day, we filtered the dataset from 36 to 32 sequences, and we release the 32 sequences. The four other sequences were omitted due to length or label errors.
Relative challenge ranking stays the same, and in this paper we report numbers for results on the 32 sequence set, STIRC2025, available here: \url{https://zenodo.org/records/20191078}.

\subsection{Useful Datasets for Tracking in Surgery}
\label{sec:datasets}

There are many datasets that can be useful for evaluating point tracking in surgery.
At MICCAI 2022, a similar challenge was organized for the same task of tracking tissue~\cite{cartuchoSurgTChallengeBenchmark2024}.
The primary differences between this method and the STIR challenge comprise our choice of a point-tracking metric~\cite{doerschTAPVidBenchmarkTracking2022}, and increased size and diversity of our data.
The STIR dataset is not labeled in a temporally dense manner, while the SurgT dataset is labelled per-frame.
For a detailed summary of useful datasets in this space, refer to the review by Schmidt et al.~\cite{schmidtTrackingMappingMedical2024a}.
Recently, some additional datasets and meta-datasets have become available.
Here is a brief list of data we recommend looking at.
Meta-MED~\cite{buddTransferringRelativeMonocular2024a} is an assembled meta-dataset.
This dataset is intended to be used for training and evaluating monocular depth models, but would also be useful for self-supervised training of tracker models.
The StereoMIS~\cite{hayozLearningHowRobustly2023} dataset comprises many stereo sequences and could be used for similar purposes.
The SurgVU dataset~\cite{ziaSurgicalVisualUnderstanding2025} also serves as a large (hundreds of hours) repository of single-eye video that could be used for self-supervised training.

\section{Dataset and Annotation}
\label{sec:dataandannotation}

\begin{figure}[t]
\centering
\includegraphics[width=\linewidth]{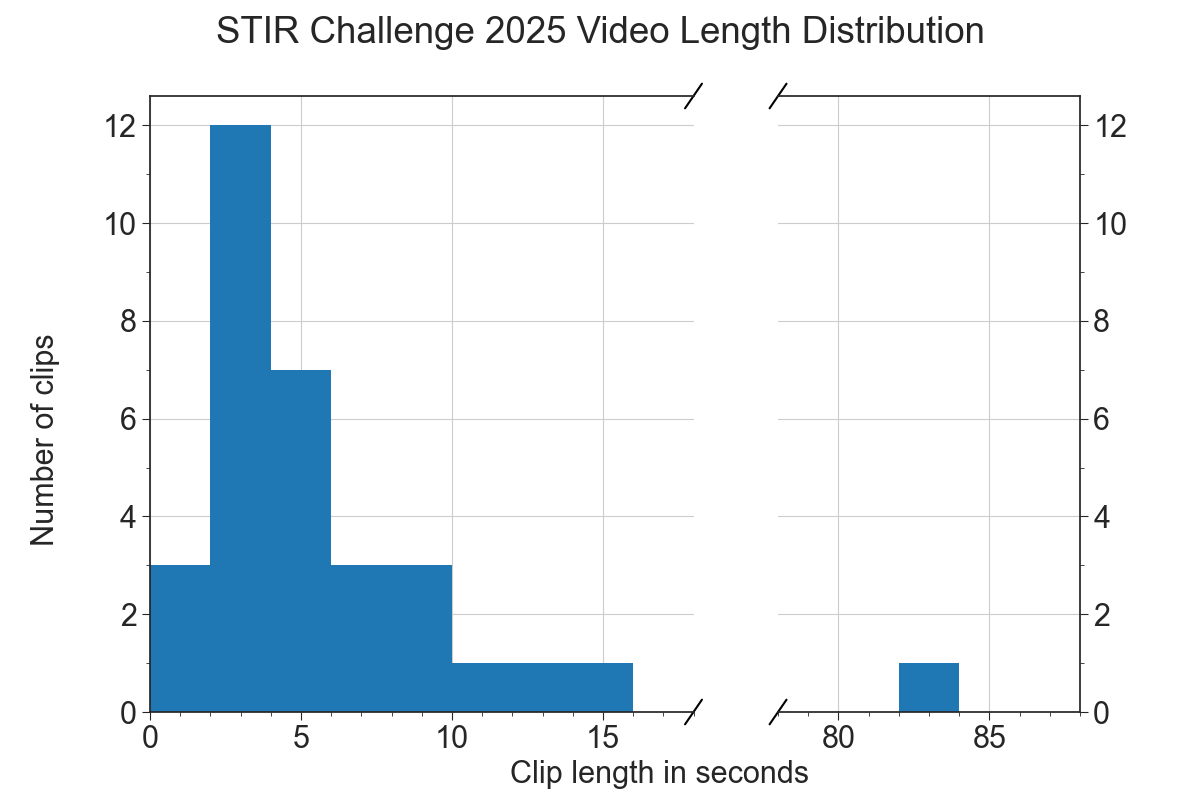}
\caption{Temporal distribution of videos. Most clips lie between 0 and 10 seconds, with a few longer clips over 20 seconds. Average clip length is 7.7 seconds.}
\label{fig:cliphistogram}
\end{figure}

\begin{figure}[t]
\centering
\includegraphics[width=\linewidth]{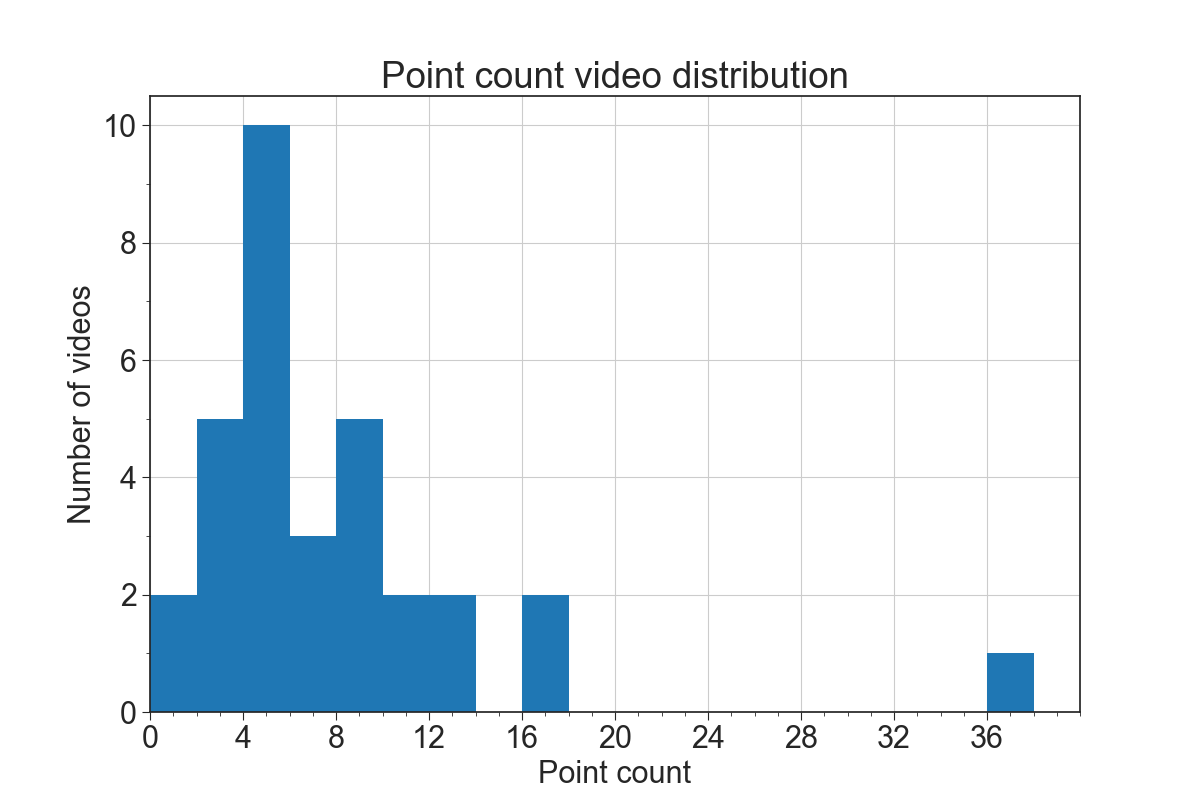}
\caption{Number of labelled points per video. Labels can be seen in Fig.~\ref{fig:startsegs}.}
\label{fig:pointhistogram}
\end{figure}

\begin{figure}[t]
\centering
\includegraphics[width=\linewidth]{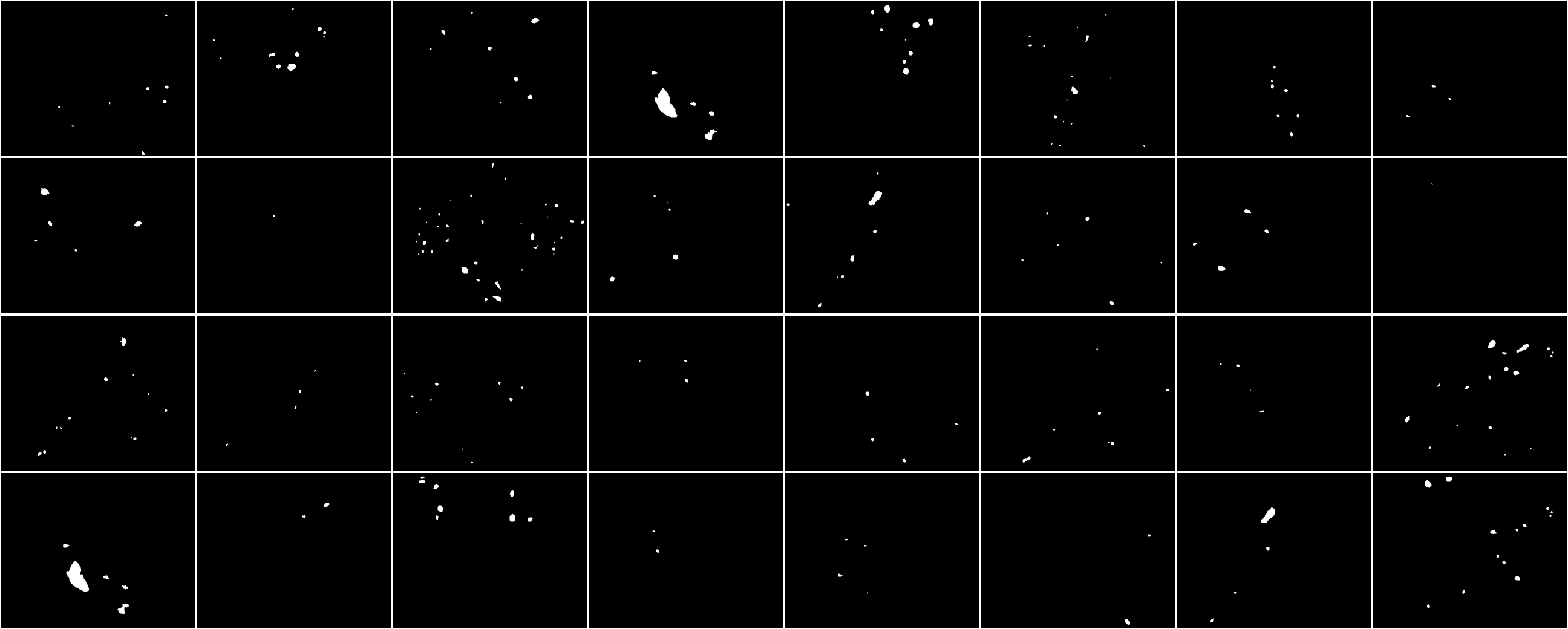}
\caption{Start point labels for all 32 sequences in in the STIR 2025 test dataset. For each sequence, center points are extracted from each segmentation, and passed to each participant's tracker.}
\label{fig:startsegs}
\end{figure}

The STIR Challenge 2025 dataset consists of sets of stereo video clips collected on a da Vinci 5 system.
Each clip consists of a start IR image and end IR image \(I_s, I_e \), segmentations of the IR ink \( S_s \) and \(S_e \), respectively, and the visible light clip, \( V \).
All frames are of size 1280 \(\times\) 1024 pixels.
\(I_s, I_e\) are in Portable Network Graphic (png) format;
\(V\) is the video clip in MPEG-4 Part 14 (mp4) format;
\(S_s, S_e\) are binary segmentations of the IR frames (png).
This dataset comprises 32 sequences.
The distribution of clip lengths is shown Fig.~\ref{fig:cliphistogram}.
Clips were on average 7.7 seconds long, with a standard deviation of 13.9 seconds.
For a histogram of the number of label points per video, refer to Fig.~\ref{fig:pointhistogram}.
No clips longer than 4 minutes are included.
Summary images of the labels can be found in Fig.~\ref{fig:startsegs}.
There are a total of 234 points over the 32 sequences.

\begin{figure*}[t]
	\centering
	\includegraphics[width=\textwidth]{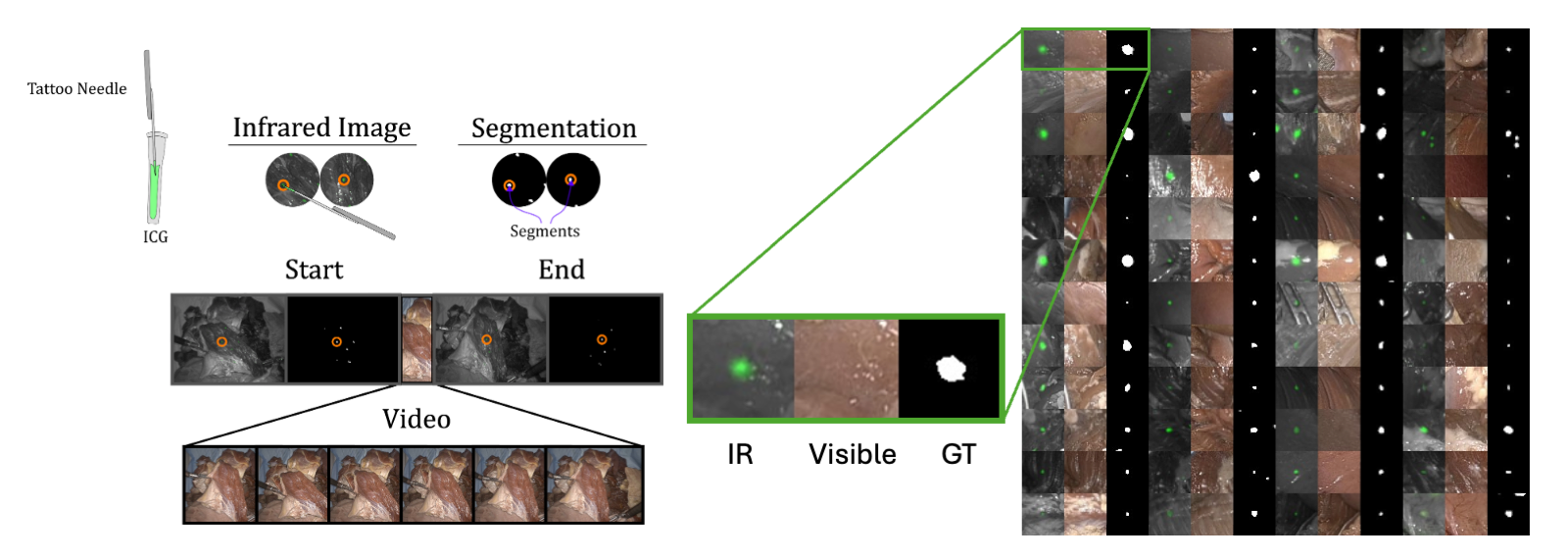}
	\caption{Dataset label creation process. Ground truth (GT) data is collected by using a tattoo needle to label points at the start and end of video frames. After tattooing is completed, multiple sequences can be collected. For each sequence, the camera captures an image in infrared (GT start frame), then switches to white light. Actions are performed under white light, and this video is recorded and saved. Then the camera switches back to IR and captures the end frame which is used as the GT for each point's motion. Segments are the binary-thresholded IR images; tattooed regions are shown in white. On the right is a figure showing a set of random triplets with the triplet: (IR image, visible light image, segment/GT image) for each point shown. Figure from~\cite{schmidtSTIRC2024}}
	\label{fig:datalabelling}
\end{figure*}

STIRC2025 was created following the same process as STIROrig~\cite{schmidtSurgicalTattoosInfrared2024}, and more detail can be found in their paper.
The labelling process is visually described in Fig.~\ref{fig:datalabelling}.
Points are tattooed with indocyanine green (ICG) ink, to create ground truth labels.
The endoscope is switched to fluorescent (IR) mode at the start and end of an action to collect the point locations at the start and end of a video.
The video for tracking is recorded in white light, and multiple actions can happen within.
The data comes from porcine subjects for the {\em in vivo} cases, and is a mix of different tissue for the {\em ex vivo} cases.

Label segmentations are calculated by first thresholding the IR-channel of the image.
An opening morphological transformation, which consists of erosion followed by dilation, is applied to reduce noise.
The resulting segments are then verified by ensuring that if a segment appears in the start image that it also appears in the end image.
To annotate, we first evaluate visibility of markers over randomly sampled cases, ensuring the tattoos do not provide features that algorithms could track~\cite{schmidtSurgicalTattoosInfrared2024}.
After this, a user looks through every case and removes noisy segmentation masks that result from specularity.
This filtering helps to reduce label noise.

In order to compute the ground-truth 3D locations, we complete an epipolar search with normalized cross correlation, using the segmented points as candidates.
This enables us to select which segment in the right image corresponds to a given segment in the left image.

The 3D position for a segment is calculated by backprojecting it.
Since the left and right eyes of the endoscope do not have the same principal point, with the left at \(c_x\) and the right at \(c'_x\), we must backproject with this in mind.
We calculate depth using the baseline \(b\), focal length \(f\), and \(c_x, c'_x\) from the calibration along with the point x-location in the left and right image (\(x, x'\)).
The depth, \(z\), is:
\[z = \frac{b*f}{(x - c_x) - (x' - c'_x)}\]

\subsection{Data Format}
\begin{figure*}[t]
	\centering
	\includegraphics[width=\textwidth]{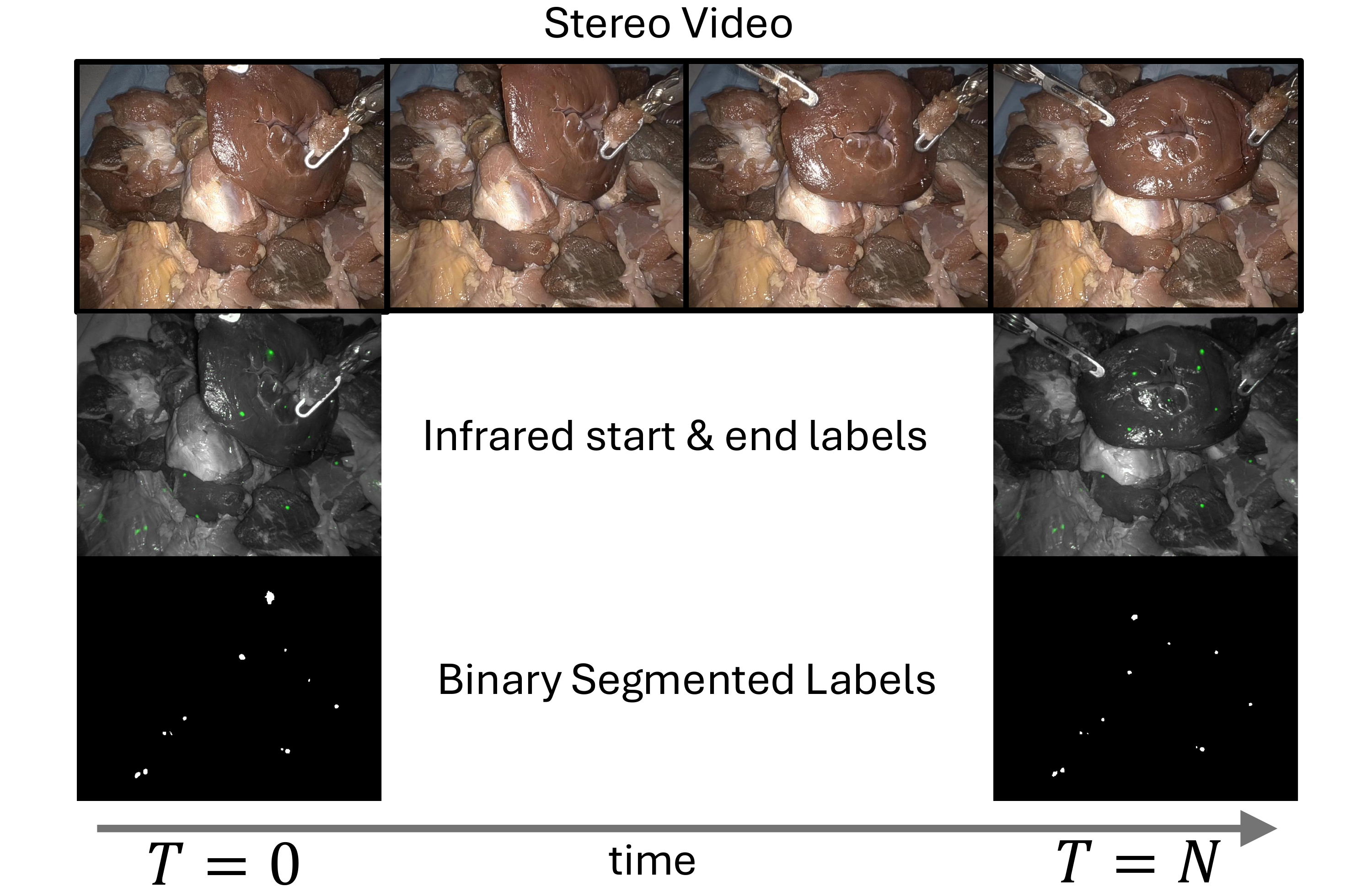}
	\caption{Example {\em ex vivo} sequence (session 06, sequence 28) from the STIR Challenge 2025 test set. The start segmentation (middle left) is converted into a set of points (bottom left), that are passed to the algorithm which tracks in the white light stereo video (top). The algorithm results are compared to the ground truth end points (bottom right).}
	\label{fig:dataexample}
\end{figure*}

We summarize the data in the STIR Challenge 2025 (STIRC2025) here, noting the format is the same as that for STIROrig~\cite{schmidtSurgicalTattoosInfrared2024} and STIRC2024~\cite{schmidtSTIRC2024}.
STIRC2025 includes a set of 8 collection sessions, named as \textbf{\code{\textless \%02d\,\textgreater}}, \((01, 02, 03, 04, 05, 06, 07, 09)\).
There are 4 {\em in vivo} sessions \((01, 03, 04, 07)\) and 4 {\em ex vivo} sessions \((02, 05, 06, 09)\). Each session includes multiple sequences.
An example sequence for one of the {\em in vivo} cases is shown in Fig.~\ref{fig:dataexample}.

\begin{itemize}
\item[] \code{left}
    \begin{itemize}
        \item[] \code{starticg.png} (Infrared image of start frame)
        \item[] \code{endicg.png} (Infrared image of end frame)
        \item[] \code{segmentation/startim.png}, 
        \item[] \code{segmentation/endim.png} (Filtered and segmented binary versions of IR start and end image)
        \item[] \code{frames/\textless ms\,\textgreater\_ms.mp4} (video file)
   \end{itemize}
\item[] \code{right}
    \begin{itemize}
        \item[] \code{starticg.png}
        \item[] \code{endicg.png}
        \item[] \code{frames/\textless ms\,\textgreater\_ms.mp4} (video file)
    \end{itemize}
\item[] \code{calib.json} Camera calibration parameters (intrinsics, relative stereo pose with translation in metres and rotation in axis-angle format)
\end{itemize}
The video file names include start and end capture times in milliseconds.


\section{Metrics}
\label{sec:metrics}

In this challenge, we evaluate the submitted algorithms based on two different metrics: accuracy and efficiency.
The accuracy metric is important for clinical verification. The efficiency metric measures the inference latency of the submitted algorithms and evaluates an algorithm's feasibility in running on clinical systems.
For the accuracy metric, we evaluate two categories: 2D trackers, and 3D trackers.

\subsection{Accuracy}

To evaluate accuracy, we use an outlier resistant metric that calculates accuracy over multiple thresholds.
This metric works for both 2D and 3D results.
The metric is \(\delta^{avg}\), introduced in TAP-Vid~\cite{doerschTAPVidBenchmarkTracking2022}, which is a non-medical point tracking challenge.
In TAP-Vid~\cite{doerschTAPVidBenchmarkTracking2022}, the points also have an occlusion score. In STIRC2025, we use data in which the points are unoccluded at the end frame.
Points can still be occluded and reappear during a sequence due to camera movement, instrument-tissue occlusion, or tissue-tissue occlusion.

To calculate the metric, \(\delta^{avg}\), each algorithm estimates the position of point(s) for each frame in a video in a streaming manner.
The estimated point(s) in the final frame, \(\hat{p}_{end}\), are used for calculating error.
The calculated finish points, \(\hat{p}_{end}\), are 2 dimensional (pixel space, left camera) for 2D trackers, and 3 dimensional for the 3D trackers.
The accuracy metric is averaged across all points and thresholds, with each point weighted evenly.
To calculate euclidean distance, each point is matched to its nearest point in the end point label set.

\begin{align}
\delta^{avg} = \Sigma_{i = i}^{M} \delta^{\mathbf{l}_i}/M\\
\delta^{\mathbf{l}_i} = \Sigma_{\hat{p}_{end}} \mathbbold{1}({d(\hat{p}_{end}, p^{nearest}_{end}) < \mathbf{l}_i}) / N
\end{align}

\(\mathbbold{1}\) is the indicator function, used to count the amount of points under the distance threshold.
\(d()\) is calculates euclidean distance in the dimension of input (2D/3D).
\(N\) is the total number of points across all videos, and \(M\) is the number of thresholds used.
Thus, \(\delta^{\mathbf{l}_i}\) is accuracy at the threshold \(\mathbf{l}_i\).
For 2D, the thresholds are \(\mathbf{l} = [4, 8, 16, 32, 64]\) with units as pixels in the full \(1024 \times 1280\) image.
For 3D, the thresholds are \(\mathbf{l} = [2, 4, 8, 16, 32]\) with units as millimetres.

\subsection{Efficiency}
The efficiency of an algorithm is measured by its computational latency, assessed across all video frames to derive a latency distribution. While the mean latency provides a general efficiency metric, it fails to capture worst-case behavior. In real-world surgical point tracking, predictability depends on worst-case and tail latencies. Therefore, we evaluate efficiency using the 95th and 99th percentile latencies in addition to the mean. The final efficiency score is the average of these three metrics, offering a comprehensive assessment of both real-time and practical performance. A submission is considered for the efficiency category only if the accuracy of the algorithm is above a certain threshold.

\section{Submissions and Baselines}
\label{sec:submissions_and_baselines}
Here, we summarize submissions to the challenge day.
We also provide the results of the baseline methods that we provided on our github page, \url{https://github.com/athaddius/STIRMetrics}.

\subsection{Baselines}
\subsubsection{MFT}
\label{subsubsection-mft}
This is the baseline MFT method~\cite{neoral2024mft}.
This method runs optical flow between a frame and multiple frames at skips into the past.
The algorithm selects its optimal trajectory by using the highest certainty unoccluded trajectory.
RAFT~\cite{teed2020raft} is used as the optical flow architecture.
In our evaluation, to maintain inference efficiency for tracking, images are downsampled by a factor of 2.
Skip factors are the same as those used in the MFT paper, \([-\infty, 1, 2, 4, 8, 16, 32]\), and the occlusion threshold is set as \(0.02\).
After tracking, locations are scaled up by 2 to get full-resolution coordinates.

\subsubsection{CSRT}
\label{subsubsection-csrt}
This method uses the Channel and Spatial Reliability tracker (CSRT~\cite{lukezicDiscriminativeCorrelationFilter2017}), initialized with one region of interest for each point in the first frame.
This tracker uses correlation based adaptive template matching to track a point across a video.
Tracking is performed on half scale images, and upscaled.
The region of interest for each point is a \(29\times 29\) box centered at the point location.

\subsubsection{RAFT}
\label{subsubsection-raft}
We use RAFT~\cite{teed2020raft} off-the-shelf to track points from one frame to the next in a streaming manner.
We track on half-scaled images, for efficiency, and multiply the final result by 2 to obtain full-resolution results.
RAFT internally iterates multiple times, refining estimates with each iteration.
We select 12 iterations for our baseline.

\subsubsection{RAFT + RAFT Stereo (3D)}
\label{subsubsection-raft3D}
This is the 3D baseline, which uses RAFT to estimate flow from one frame to the next in the left eye, and finds the 3D position by backprojecting the points using the disparity estimate and the known camera parameters.
The disparity estimate is calculated using RAFT-Stereo~\cite{lipson2021raft}.

\subsubsection{Control}
\label{subsubsection-control}

The baseline control method estimates 0 motion for every point.
This provides a minimum bound of accuracy, which is useful for debugging.
This also allows the organisers and participants to ensure their methods use the correct data.
The control method runs alongside submissions.
During the challenge, this serves as a useful sanity check.

\subsection{Submissions}
\label{submissions}

Seven teams submitted reports for the 2025 challenge. All seven of them participated in the 2D component, and two also participated in the 3D tracking tasks.

\subsubsection{BlueJays}
\label{BlueJays}

\begin{figure}[tbp]
\centering
\includegraphics[width=\linewidth]{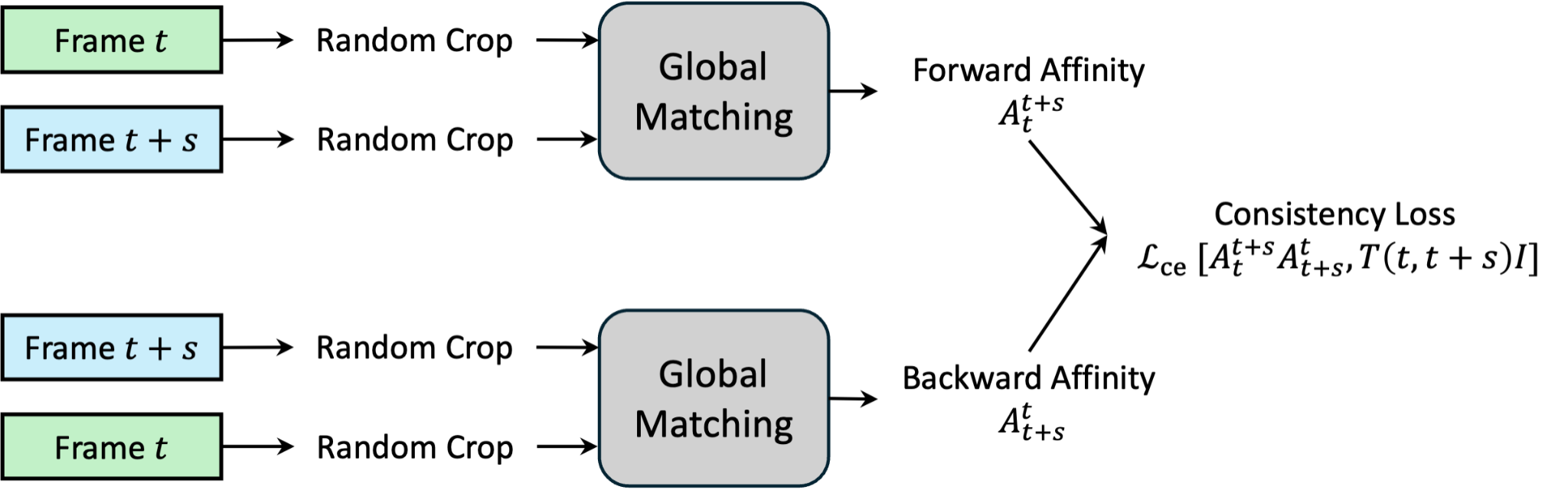}
\caption{Overview of the solution and the consistency-based self-supervised training method provided by team BlueJays.}
\label{BlueJays-main-fig}
\end{figure}

Team BlueJays (Fig.~\ref{BlueJays-main-fig}) proposes a self-supervised tissue tracking method based on contrastive random walks and global matching. The method formulates point tracking as a random walk over a space-time graph, where image patches or pixels are nodes and learned affinities between adjacent frames define the graph edges. A GMFlow~\cite{xu2022gmflow} global matching transformer is fine-tuned to produce pixel-wise correspondences, supervised by cycle consistency over frame pairs. To regularize the learned correspondences, the method additionally applies an edge-aware smoothness loss.

For inference, BlueJays uses a moving reference frame strategy to handle occlusion. Points are propagated frame by frame, and visibility is estimated using a forward-backward consistency check. If the end-point error exceeds a threshold, the method retries using a cached frame corresponding to the last frame where the point was visible. If the point remains inconsistent with both the reference and cached frame, it is considered occluded and its flow update is skipped. The model is initialized from official GMFlow pretrained weights and trained on STIROrig~\cite{schmidtSurgicalTattoosInfrared2024} and SurgT~\cite{cartuchoSurgTChallengeBenchmark2024} using 720-by-720 padded frames with random 640-by-640 crops.

\subsubsection{CCG\_DGIST}
\label{CCG_DGIST}

\begin{figure}[tbp]
\centering
\includegraphics[width=\linewidth]{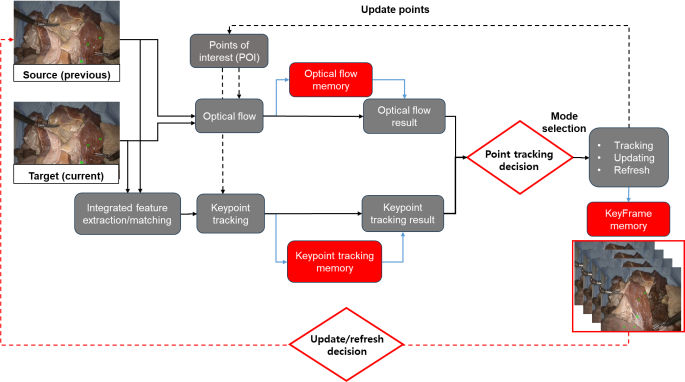}
\caption{Overview of the two-branch solution that combines optical flow and sparse keypoint matching provided by team CCG\_DGIST.}
\label{ccg-dgist-main-fig}
\end{figure}

Team CCG\_DGIST (Fig.~\ref{ccg-dgist-main-fig}) proposes a hybrid 2D tracking framework that combines dense optical flow with keypoint-based tracking. The method exploits the complementary strengths of the two approaches: optical flow provides accurate local motion estimates, while keypoint-based tracking improves long-term robustness under drift, occlusion, and large deformation.

The pipeline runs optical-flow and keypoint-based modules in parallel. The keypoint branch combines SuperPoint~\cite{detone2018superpoint}, ALIKED~\cite{zhao2023aliked}, LightGlue~\cite{lindenberger2023lightglue}, and RoMA~\cite{edstedt2024roma} to obtain reliable feature correspondences, while the optical-flow branch estimates fine-grained point motion. The distance between the two predictions is used as a confidence measure to switch between three modes. In tracking mode, the optical-flow estimate is used when both predictions agree. In update mode, points with large disagreement are temporarily frozen and later recalibrated from keyframe-based estimates. In refresh mode, point positions are periodically updated from keyframe references to reduce accumulated drift. In the reported implementation, keyframes are selected every three frames and stored in a rolling buffer of fifteen frames.

\subsubsection{MFTIQProb}
\label{MFTIQProb}

\begin{figure}[tbp]
\centering
\includegraphics[width=\linewidth]{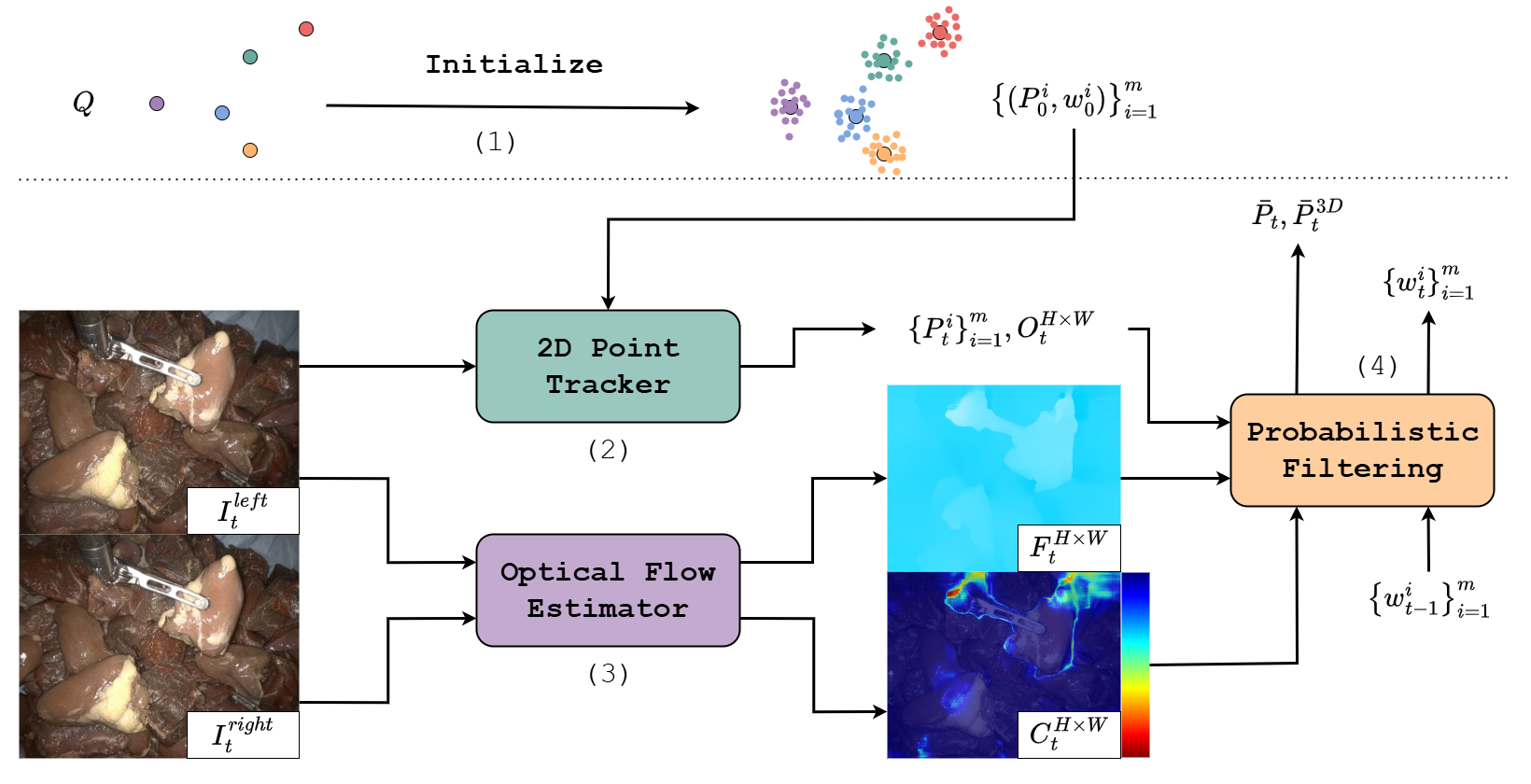}
\caption{Overview of the solution for 2D and 3D tracking provided by team MFTIQProb.}
\label{MFTIQProb-main-fig}
\end{figure}

Team MFTIQProb (Fig.~\ref{MFTIQProb-main-fig}) proposes a probabilistic multi-point tracking pipeline for both 2D and 3D tracking. Instead of representing each query as a single point, the method initializes each track with a Gaussian distribution of candidate points around the query location. Each candidate is assigned an initial likelihood weight, creating an prior over plausible local surface motion.

For 2D tracking, the team uses an off-the-shelf point tracker, MFT~\cite{neoral2024mft} or MFTIQ~\cite{serych2024mftiq}, to estimate per-frame point positions and visibility scores. For 3D tracking, the method estimates bidirectional stereo flow using Sea-RAFT~\cite{wang2024sea} and backprojects the resulting stereo correspondences using the provided calibration matrix. Forward and backward stereo estimates are compared in 3D, and the forward-backward discrepancy is converted into a Gaussian likelihood for each candidate. A normalized Bayesian update is then applied when the candidate is visible; occluded candidates retain their previous weights until reobserved. The final 2D and 3D predictions are computed as weighted averages over the candidate distribution.

\subsubsection{MoriLabNU}
\label{MoriLabNU}

Team MoriLabNU adapts the MFT~\cite{neoral2024mft} to laparoscopic point tracking using Elastic Weight Consolidation (EWC). The approach fine-tunes the convolutional layers of the MFT feature encoder while freezing the flow chaining and occlusion reasoning components. This preserves MFT’s long-term tracking and occlusion-handling behavior while allowing the feature extractor to adapt to surgical textures, lighting changes, specularities, and deformable tissue.

To reduce catastrophic forgetting, the adaptation objective includes an EWC penalty based on the diagonal Fisher Information matrix estimated from source STIROrig~\cite{schmidtSurgicalTattoosInfrared2024} sequences. Since dense target annotations are unavailable, the method uses a frozen MFT teacher to generate pseudo-label trajectories. These trajectories are filtered using a backward tracking consistency check, and the retained tracks supervise the adapted encoder with a Huber loss. The pipeline also applies CLAHE to the value channel in HSV color space to improve local contrast, and adjusts MFT hyperparameters by expanding temporal delta intervals, increasing the occlusion threshold, and increasing flow iterations for soft-tissue deformation.

\subsubsection{NCT\_TSO}
\label{NCT_TSO}

\begin{figure}[tbp]
\centering
\includegraphics[width=\linewidth]{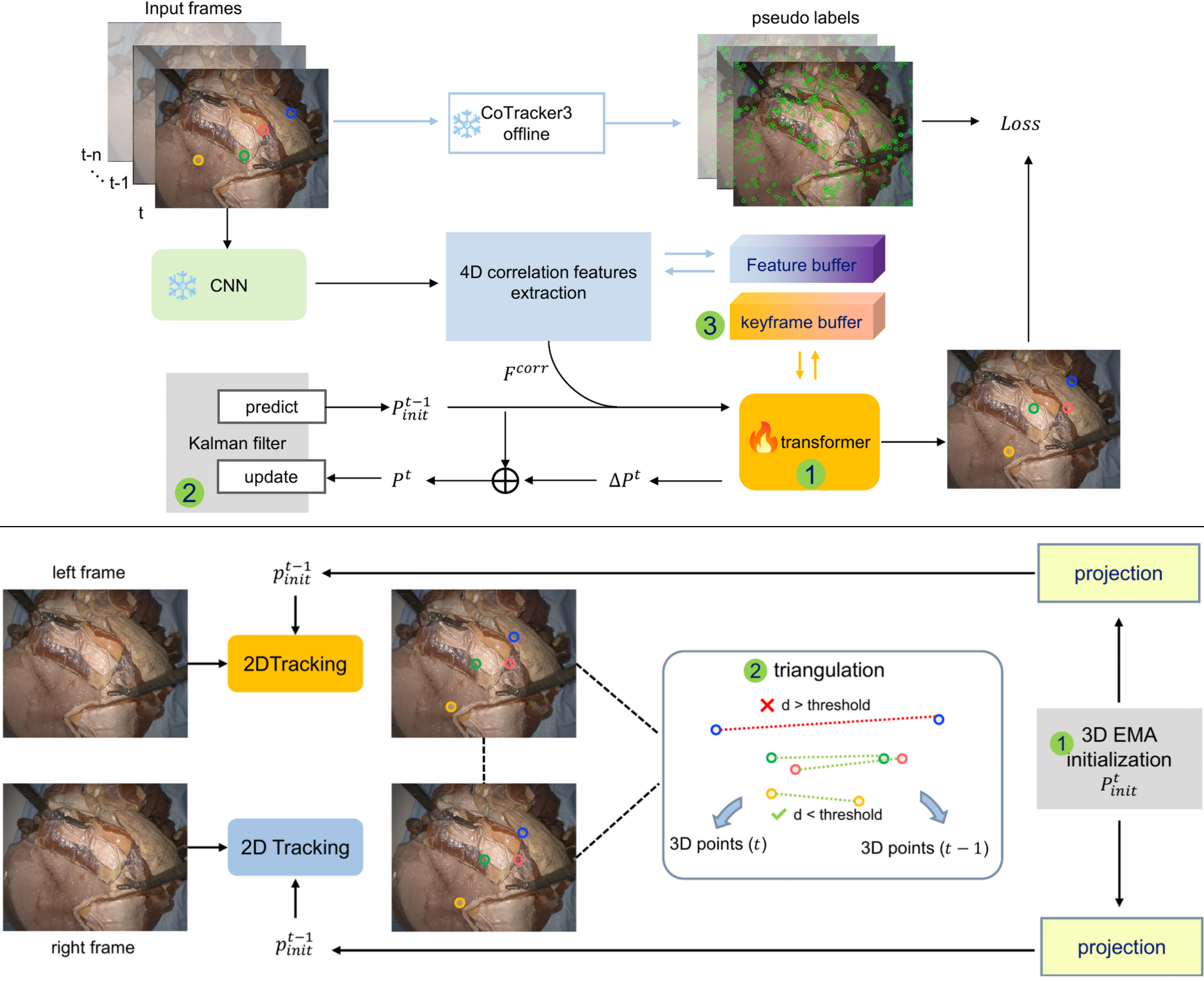}
\caption{Overview of the solution for 2D (above) and 3D (below) tracking tasks by team NCT\_TSO.}
\label{NCT_TSO-main-fig}
\end{figure}

Team NCT\_TSO (Fig.~\ref{NCT_TSO-main-fig}) builds a 2D tracking pipeline on LiteTracker~\cite{karaoglu2025litetracker}, a real-time tracker derived from CoTracker3~\cite{karaev2024cotracker3}. LiteTracker accelerates online tracking using a ring buffer for cached 4D correlation features and an exponential moving average flow initialization. The submitted method starts from the official pretrained online CoTracker3 weights and fine-tunes the tracker on STIR~\cite{schmidtSurgicalTattoosInfrared2024} clips using pseudo-ground-truth trajectories generated by the stronger offline version of CoTracker3. During training, the feature extraction CNN is frozen and only the transformer-based update module is trained with a Huber loss.

The team modifies LiteTracker in two ways. First, for 2D tracking, the original EMA coordinate initializer is replaced with a linear Kalman filter that models each point using position and velocity, allowing the tracker to balance motion-model predictions with incoming measurements according to process and measurement uncertainty. Second, a keyframe feature buffer is introduced to reduce drift caused by large tissue deformation. High-confidence frames are stored as keyframes, and every 20 frames their features are reused by the transformer for an additional refinement step. For 3D tracking, two finetuned LiteTrackers run in parallel on the left and right stereo streams. The resulting 2D points are triangulated using the provided calibration, while a 3D EMA motion model reprojects predicted 3D positions back into both image planes to initialize the next 2D tracking step. A 3D geometric consistency check rejects triangulated measurements that deviate too far from the motion-model prediction.

\subsubsection{SRV}
\label{SRV}

Team SRV proposes MFT-WAFT, a variant of the MFT~\cite{neoral2024mft} that replaces the original RAFT optical-flow component with WAFT~\cite{wang2025waft}, a recently introduced optical-flow method based on warping-alone field transforms. 

WAFT differs from RAFT~\cite{teed2020raft} by avoiding large all-pairs cost volumes. Instead, it starts from an initial motion estimate, warps features across frames, compares the corresponding features, and iteratively refines the optical-flow prediction. In the submitted pipeline, this WAFT flow output is inserted into the standard MFT framework. MFT then propagates pixel locations through time using the WAFT flow fields, while retaining its original multi-interval temporal skips and occlusion- and uncertainty-aware reasoning. 

\subsubsection{UII\_VRI}
\label{UII_VRI}

\begin{figure}[tbp]
\centering
\includegraphics[width=\linewidth]{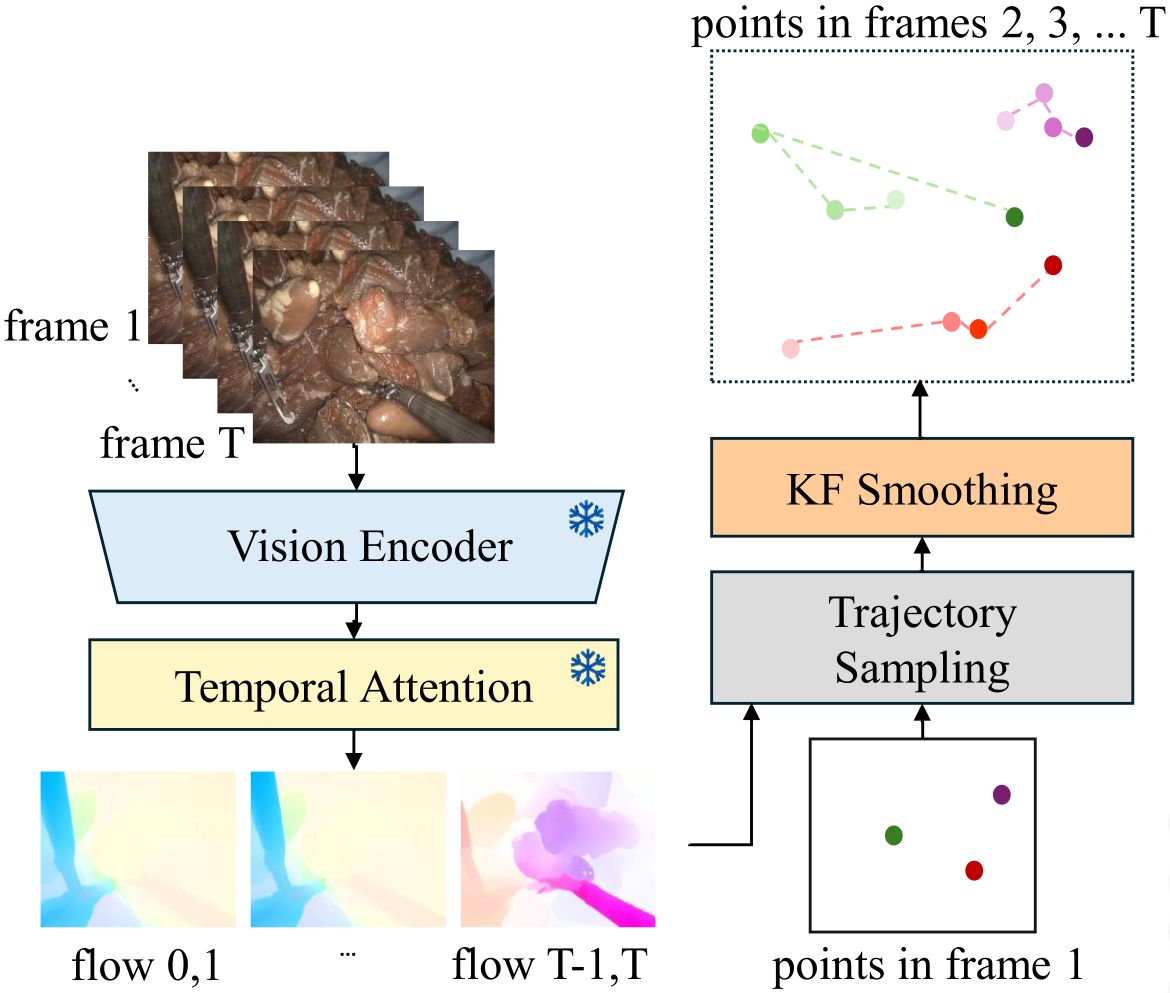}
\caption{Overview of the solution provided by team UII\_VRI.}
\label{UII_VRI-main-fig}
\end{figure}

Team UII\_VRI (Fig.~\ref{UII_VRI-main-fig}) proposes SurgTrack, a long-range 2D point tracking pipeline that does not rely on surgical-domain fine-tuning. The method uses a frozen, dense point-tracking model, AllTracker~\cite{harley2025alltracker}, pretrained on general non-medical data. A vision encoder extracts frame-wise features over a temporal window, and a temporal attention module estimates dense correspondences from the first frame to subsequent frames. Query point trajectories are then sampled from these correspondence fields.

To reduce temporal jitter, SurgTrack applies Kalman smoothing as a post-processing step to each predicted trajectory. Each point is modeled with a constant-velocity state consisting of 2D position and velocity, and the neural tracker output is treated as a noisy measurement. The submitted implementation processes STIR videos at native 1280-by-1024 resolution  and computes frame features on the fly over windows of 16 frames.

\section{Results}
\label{sec:results}

This section will summarize the results of each participating team.

\subsection{Accuracy}

\begin{table}[tbp]
\caption{2D tracking accuracy comparison. $\delta^{\mathbf{l}_i}$ indicates the tracking accuracy where $\mathbf{l}_i$ is the threshold defined in pixels.}
\label{tab:accuracy-2d}
\begin{tabularx}{\columnwidth}{X c c c c c c c}
    \toprule
    \multirow{2}{*}{\textbf{Method}} & \multicolumn{6}{c}{$\delta^{\mathbf{l}_i}\uparrow$}\\
    \cmidrule{2-7}
    & $\mathbf{l}_i=4$ & $=8$ & $=16$ & $=32$ & $=64$ & Avg. \\
    \midrule
    \multicolumn{7}{X}{\textbf{Baselines}} \\
    \midrule
    CONTROL & 17.95 & 38.03 & 53.42 & 62.39 & 78.21 & 50.00\\
    RAFT & 9.40 & 24.79 & 58.12 & 80.77 & 89.74 & 52.56\\
    CSRT & 26.50 & 61.97 & 80.34 & 85.04 & 88.46 & 68.46\\
    MFT & 48.29 & 75.64 & 95.30 & 98.72 & 99.57 & 83.50\\
    \midrule
    \multicolumn{7}{X}{\textbf{2025 Subm.}} \\
    \midrule
    SRV & 16.67 & 41.88 & 72.22 & 85.47 & 90.17 & 61.28\\
    BlueJay & 34.19 & 61.97 & 81.62 & 87.18 & 91.03 & 71.20\\
    NCT\_TSO & 43.59 & 72.65 & 91.88 & 96.15 & 98.72 & 80.60\\
    UII-VRI-2D & 46.15 & 76.50 & 94.87 & 97.44 & 98.29 & 82.65\\
    MoriLabNU & 44.87 & 76.50 & 94.87 & 98.29 & 99.15 & 82.74\\
    MFTIQProb & 45.73 & 74.36 & 95.73 & 99.15 & 99.57 & 82.91\\
    CCG\_DGIST & 47.01 & 74.79 & 94.87 & 98.72 & 99.57 & 82.99\\
    \bottomrule
\end{tabularx}
\end{table}

\begin{table}[tbp]
\caption{3D tracking accuracy comparison. $\delta^{\mathbf{l}_i}$ indicates the tracking accuracy where $\mathbf{l}_i$ is the threshold defined in millimetres.}
\label{tab:accuracy-3d}
\begin{tabularx}{\columnwidth}{X c c c c c c c}
    \toprule
    \multirow{2}{*}{\textbf{Method}} & \multicolumn{5}{c}{$\delta^{\mathbf{l}_i}\uparrow$}\\
    \cmidrule{2-7}
    & $\mathbf{l}_i=2$ & $=4$ & $=8$ & $=16$ & $=32$ & Avg. \\
    \midrule
    \multicolumn{7}{X}{\textbf{Baselines}} \\
    \midrule
    CONTROL & 19.82 & 40.53 & 59.91 & 79.30 & 93.83 & 58.68\\
    RAFT\_Stereo & 17.95 & 40.60 & 60.26 & 82.48 & 92.74 & 58.80\\
    \midrule
    \multicolumn{7}{X}{\textbf{2025 Subm.}} \\
    \midrule
    MFTIQProb & 18.38 & 38.03 & 57.26 & 78.63 & 93.16 & 57.09\\
    NCT\_TSO & 33.04 & 51.54 & 75.33 & 87.67 & 94.71 & 68.46\\
    \bottomrule
\end{tabularx}
\end{table}

Table~\ref{tab:accuracy-2d} provides the overarching summary for 2D methods.
Of the ranked challenge submissions, Team CCG\_DGIST came first, with a \(\delta^{avg} = 82.99\).
Team MFTIQProb was second with a \(\delta^{avg} = 82.91\), and Team MoriLabNU was third with a \(\delta^{avg} = 82.74\).
The overall best performing method was MFT~\cite{neoral2024mft} with \(\delta^{avg} = 83.50\).

Table~\ref{tab:accuracy-3d} summarizes the 3D submissions.
Of the challenge submissions, Team NCT\_TSO came in first with a \(\delta^{avg} = 68.46\), and Team MFTIQProb came in second with a \(\delta^{avg} = 57.09\).

\subsection{Efficiency}
In this section we summarize the efficiency results.
All the teams except CCG\_DGIST participated in the this component of the challenge.
The efficiency of a method is computed by the 95th percentile per-frame latency across all frames in all sequences that are used in the 2D tracking challenge.
To standardize this, we ran all the submissions on RTX A5000 GPU.

To strike a meaningful balance between tracking accuracy and efficiency, we set a threshold to discard methods that achieve less than 90\% of the accuracy of the top-performing method.
Consequently, the solutions provided by teams SRV~\ref{SRV} and BlueJay~\ref{BlueJays} fell short of this threshold and were omitted from the evaluation.

The efficiency challenge winner, NCT\_TSO~\ref{NCT_TSO}, achieved a per-frame latency of $26.25$ ms while performing comparably to the most accurate method.
This latency measurement is within the acceptable range for many surgical point-tracking applications in addition to being significantly lower than the closest competitor, UII\_VRI~\ref{UII_VRI}, which achieved $ 140.52 $ ms.

\begin{figure}[tbp]
\centering
\includegraphics[width=0.9\linewidth]{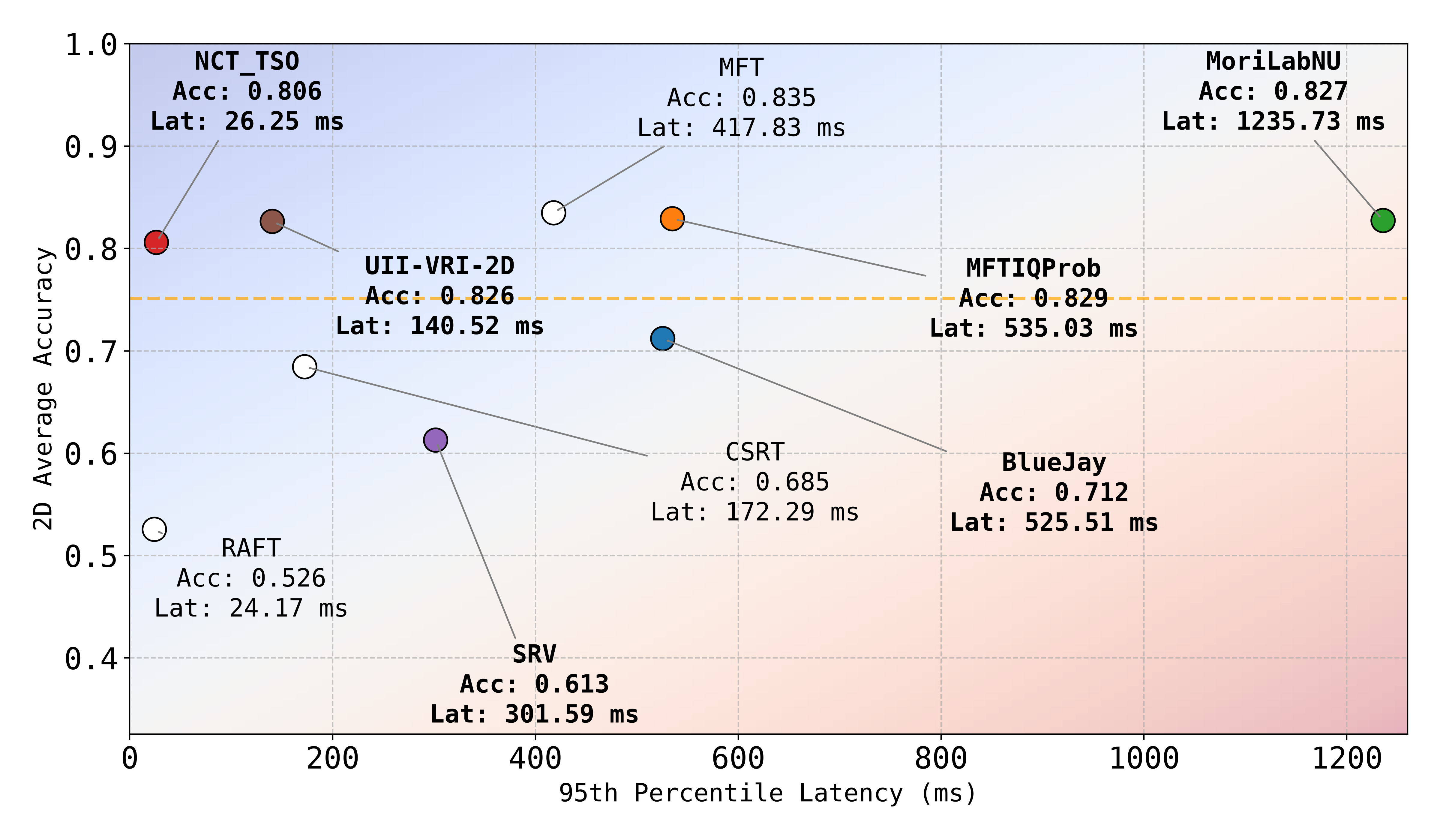}
\caption{Latency and 2D accuracy comparisons for efficiency evaluation. Baseline methods are indicated with white, submissions are indicated with colored markers. Latencies are computed as the 95th percentile of all per-frame latencies on the same workstation. The orange line is threshold set as the 90\% of the highest-accuracy of the top-method, setting the entrance bar for the efficiency challenge.}
\label{fig:latency-vs-2d-acc}
\end{figure}

\section{Discussion}
\label{sec:discussion}

In this section, we discuss the challenge results in Section~\ref{sec:accandeff}, and close with recommended focus areas in Section~\ref{sec:algodirections}.

\subsection{Accuracy and Efficiency}
\label{sec:accandeff}

Here we discuss the 2D results, followed by the 3D results and efficiency.

Again, in 2025 the most accurate method was the baseline MFT~\cite{neoral2024mft}.
Among submissions, MFT adaptations (MFTIQProb, MoriLabNU) also held the 2nd and 3rd space, with CCG\_DGIST who use a classical method combined with modern components occupying the 1st place.
We believe these results are due to it being easy to over-validate on the test+validation sets we release each year. A baseline having the strongest result emphasizes the importance of proper cross-validation and generalization in submissions.

For 3D methods, Team NCT\_TSO had the most accurate submission, beating all baselines, and team MFTIQProb came second.
We believe the 3D Kalman filtering and consistency check are why NCT\_TSO did well, since neither the baseline nor MFTIQProb used this same approach.
Of note, we have yet to see a method that tracks fully in 3D, rather than methods which use a filtered combination of 2D + stereo.
Architectural complexity could be a large factor here.

Finally, for the efficiency component, team NCT\_TSO won by using an adaptation of LiteTracker~\cite{karaoglu2025litetracker} which is an efficient adaptation of CoTracker3~\cite{karaev2024cotracker3}. Even the next-fastest method to NCT\_TSO's had over 5x the latency of NCT\_TSO's submission. As methods increase in accuracy, we expect them to increase in runtime; having a category for efficiency motivates practical solutions that are feasible for edge deployment.

\subsection{Algorithmic Direction}
\label{sec:algodirections}

We continue to encourage future methods to focus on the following:
\begin{itemize}
    \item Robust cross-validation and evaluation using small amounts of data.
    \item Pseudo-labelling, data augmentation, student teacher models applied to surgical scenarios (ie. methods like CoTracker3~\cite{karaev2024cotracker3}, BootsTap~\cite{doersch2024bootstap})
    \item Using pretrained models with surgical image features (masked autoencoders, self-supervision, etc.)
    \item Using stereo data to improve tracking.
    \item Designing low latency, efficient models for long-term tracking, and relocalization.
\end{itemize}

\section{Conclusion}
\label{sec:conclusion}

In this paper, we summarize and discuss the design, methods, and results from the 2025 STIR Challenge, which was organized as a part of EndoVis at MICCAI 2025.
This challenge alongside the publicly released test dataset serves as a resource for methods to test, compare, and iterate on algorithms for tissue tracking and other applications in robotic assisted surgery.

\bibliographystyle{splncs04}
\bibliography{reflist}

\end{document}